# Improving Panoptic Segmentation for Nighttime or Low-Illumination Urban Driving Scenes

Ankur Chrungoo
*17-Aug-2022*
Supervisor: Muhammad Sami Siddiqui
MSc in Artificial Intelligence, Queen
Mary University of London

*Abstract*— Autonomous vehicles and driving systems use scene parsing as an essential tool to understand the surrounding environment. Panoptic segmentation is a state-of-the-art technique which proves to be pivotal in this use case. Deep learning-based architectures have been utilized for effective and efficient Panoptic Segmentation in recent times. However, when it comes to adverse conditions like dark scenes with poor illumination or nighttime images, existing methods perform poorly in comparison to daytime images. One of the main factors for poor results is the lack of sufficient and accurately annotated nighttime images for urban driving scenes. In this work, we propose two new methods, first (Approach-1) to improve the performance, and second (Approach-2) to improve the robustness of panoptic segmentation in nighttime or poor illumination urban driving scenes using a domain translation approach. The proposed approach makes use of CycleGAN (Zhu *et al.*, 2017) to translate daytime images with existing panoptic annotations into nighttime images, which are then utilized to retrain a panoptic segmentation model to improve performance and robustness under poor illumination and nighttime conditions. In our experiments, Approach-1 demonstrates a significant improvement in the panoptic segmentation performance on the converted Cityscapes dataset with more than +10% PQ, +12% RQ, +2% SQ, +14% mIoU and +10% AP50 absolute gain. Approach-2 demonstrates improved robustness to varied nighttime driving environments. Both the approaches are supported via comprehensive quantitative and qualitative analysis.

## I. Introduction

Humans possess an innate ability to comprehend complex visual scenes from an early age, which enables them to interact with their environment to achieve their goals intelligently. Similarly, autonomous systems like self-driving vehicles need a holistic scene understanding to produce intelligent and safe behaviour. Since autonomous vehicles have been very popular in recent years, much research has been conducted on the urban driving scene understanding.

A visual scene or image can be divided into 'stuff' and 'thing' types of objects. 'Stuff' is defined as the background, uncountable or amorphous parts of an image such as road, sidewalk, and sky, while 'thing' represents the foreground, countable objects such as vehicles, pedestrians, and riders. Objects of 'Stuff' classes are primarily segmented through semantic segmentation, while objects of 'thing' classes are mainly segmented using instance segmentation. More specifically, in driving scenes, semantic segmentation helps in segmenting semantically different objects like cars, roads, and the sky. In contrast, instance segmentation provides object instance level masks, thus separating individual objects of each class, such as each car or pedestrian. While there have been numerous studies on these two segmentation techniques individually, panoptic segmentation was introduced by (Kirillov *et al.*, 2019) as a unified segmentation task that generates rich, coherent, and complete scene segmentation. It provides the ability to comprehend visual scenes at both the pixel level and object instance level.

With the recent advances in deep learning-based panoptic segmentation of urban driving scenes, the performance has significantly improved for daytime images due to the availability of labelled data in large driving datasets such as Cityscapes (Cordts *et al.*, 2016), Indian driving dataset (IDD) (Varma *et al.*, 2019), Mapillary Vistas (Neuhold *et al.*, 2017). However, in low illumination conditions, the quality of the segmentation task degrades significantly. This happens due to factors such as under/over-exposure, motion blur, and noise in the nighttime images. The lack of contrast between background and foreground objects also increases uncertainty in the object boundaries. Due to the above reasons, the features extracted by the convolutional layers under poor illumination conditions are much different from those extracted from daytime datasets. Thus, it leads to poor performance when models trained heavily on daytime images are used for inference on the nighttime images.

However, we cannot expect to avoid nighttime conditions in real-world applications of panoptic segmentation such as self-driving vehicles. Even then, till the end of this study, there had been no research specifically on handling nighttime images for the panoptic segmentation task, including the urban driving scenarios. One of the primary reasons for this is the lack of high-quality labelled nighttime datasets with panoptic annotations. Due to over/under exposure and indiscernible regions in nighttime driving images, it is often difficult to manually annotate and build high-quality pixel-level annotations as ground truth for training deep learning models. Thus, there is a lack of large nighttime urban driving datasets with panoptic annotations. Several existing studies in semantic segmentation have applied the domain translation approach to train models for nighttime images by making use of the daytime datasets and the respective annotations that are readily available. However, to our knowledge, the domain translation approach has never been used to solve the panoptic segmentation task for nighttime or low illumination conditions.

The following sections will provide the problem statement and the objectives (II), the contributions made (III), the background literature (IV), the methodology used (V), experiments conducted (VI), results and discussion (VII), the conclusion (VIII) and some possible future works (XI).

## II. Problem Statement

In this research, we aim to improve panoptic segmentation for nighttime or low illumination urban driving scenes. Thus, the primary research questions are formulated as below:

- Can we improve the performance of panoptic segmentation on nighttime or poorly illuminated urban driving scenes? Can it be improved via domain adaptation of existing daytime urban driving datasets?
- Can the approach be made more robust to a variety of urban nighttime driving environments?

Based on these, the following high-level objectives were outlined to conduct the research:

- Explore the possibilities of using domain adaptation model architectures to translate daytime driving scenes to nighttime or bad illumination images.
- Utilize different nighttime datasets to improve robustness of the method under varied nighttime driving conditions.
- Perform quantitative and qualitative studies to ascertain the performance improvement, if any.

### III. CONTRIBUTION

The main contributions of this research are summarized as follows:

- We propose a new method for improving panoptic segmentation of urban driving scenes under nighttime or poor illumination conditions.
- The work demonstrates that the domain adaptation technique can be used and is effective in improving panoptic segmentation in nighttime/low illumination urban driving scenes.
- It also demonstrates that this improvement can be made while still maintaining the performance for the daytime images.
- We propose a method to improve robustness of panoptic segmentation under varied nighttime urban driving environments having different illumination patterns.
- The study presents quantitative metrics and qualitative results for panoptic segmentation under nighttime urban driving scenes to substantiate the improvements.

### IV. BACKGROUND

#### A. Panoptic Segmentation

Panoptic segmentation is a unified task of segmenting both 'thing' and 'stuff' in images that was introduced and formulated by (Kirillov *et al.*, 2019) along with a metric called Panoptic Quality (PQ) that captures the performance of all 'stuff' and 'thing' classes in a unified manner. More precisely, if a pixel in an image corresponds to the 'stuff' class, the model assigns a label from one of the 'stuff' classes, whereas if the pixel corresponds to the 'thing' class, the model predicts not only the 'thing' class it belongs to but also the specific instance of the object. The panoptic segmentation task is typically addressed by separate semantic and instance segmentation networks or a single network with separate heads for each sub-task. The results of both the sub-tasks are combined to develop a panoptic segmentation result using some heuristics or post-processing. We can broadly group panoptic segmentation approaches into two different categories: top-down methods and bottom-up methods. Most of the recent methods developed in deep learning take the top-down approach for panoptic segmentation. Panoptic FPN (Kirillov and Doll, 2019), UPSNet (Xiong, Liao and Zhao, 2019), AUNet (Li *et al.*, 2019) and OCFusion (Lazarow *et al.*, 2020) are some of the seminal studies for improvements in the top-down approach. However, the top-down methods have an inherent problem due to overlaps between the predictions of both heads. DeeperLab (Yang *et al.*, 2019), Panoptic-DeepLab (Cheng *et al.*, 2020) and Axial-DeepLab (Wang *et al.*, 2020) are some of the most significant bottom-up models for panoptic segmentation. However, these bottom-up model architectures mainly suffer in performance when used on highly deformable objects. In a recent branch of study, some single path methods have been proposed to predict the results on things and stuff classes directly, out of which the most prominent ones are Panoptic FCN (Li *et al.*, 2021), Max-DeepLab (Wang *et al.*, 2021), and K-Net (Pang, Chen and Loy, 2021). SPINet (Hwang and Kim, 2022) is another single path method that tries to combine the advantages of both top-down and bottom-up approaches.

We have used the Panoptic-DeepLab (Cheng *et al.*, 2020) model architecture for panoptic segmentation in this work.

#### B. Domain adaptation

Domain adaptation methods in deep learning are very popular to transfer learning from a source domain to a target domain, where both domains have similar objects but different distributions of data. This is a key technique used when the source domain has richly labelled data but the target domain has unlabelled data. Since (Goodfellow *et al.*, 2014) proposed Generative Adversarial Networks (GAN), it has become one of the most widely used methods for image stylization. Pix2Pix (Isola *et al.*, 2017) is a very popular GAN for image-to-image translation but needs X/Y image pairs for performing the task. A recently introduced CycleGAN (Zhu *et al.*, 2017) can perform a complete translation cycle with unpaired images of different domains without needing image pairs. The goal of CycleGAN is to learn a mapping G: X →Y so that the image distribution from G(X) cannot be distinguished from the Y distribution and learn an inverse mapping F: Y → X so that F(G(X)) is equivalent to X and vice versa, via a cycle consistency loss.

In this work, we have used CycleGAN for domain translation since it can perform the task without needing paired images.

#### C. Nighttime segmentation / scene parsing

Many researchers have used the idea of domain adaptation for low illumination/nighttime semantic segmentation tasks. (Dai and Gool, 2018) aimed at reducing the cost of human annotation for nighttime images by a progressive domain adaptation. They adapted semantic segmentation models trained on daytime images to the nighttime domain via an intermediate twilight domain and also compiled the Nighttime driving dataset (Dai and Gool, 2018). (Sakaridis et al., 2019) developed a curriculum framework for day-to-night adaptation of semantic segmentation models using unlabelled real images and labelled synthetic images, and introduced the Dark Zurich dataset (Sakaridis et al., 2019) having twilight-day-nighttime correspondences for this purpose. (Nag, Adak and Das, 2019) proposed a deep learning architecture NiSeNet that uses domain mapping from synthetic to real data to semantically segment nighttime images and introduced the Urban Night Driving Dataset (UNDD) in the process. (Romera *et al.*, 2019) and (Sun *et al.*, 2019) proposed two similar approaches for semantic segmentation where domain adaptation was used to convert daytime images to nighttime for training with the daytime annotations, and nighttime images were converted to more easily processable daytime images at the inference time. While the former applied the UNIT (Liu, Breuel and Kautz, 2017) framework for day-to-night image translation, the latter employed CycleGAN (Zhu *et al.*, 2017) for translating daytime images to nighttime. (Wu *et al.*, 2021) recently proposed a novel end-to-end deep learning model DANNet for semantic segmentation of nighttime images in an unsupervised manner. DANNet uses a multi-target domain adaptation via adversarial learning to handle the semantic segmentation in nighttime images, which internally uses an image relighting network, a semantic segmentation network and two discriminators. It is the first one-stage domain adaptation model proposed for semantic segmentation of nighttime images that does not need a separate training of a GAN (Goodfellow *et al.*, 2014) for domain translation from daytime to nighttime or vice versa.

(Tan *et al.*, 2021) introduced an exposure-aware method which is end-to-end trained and makes use of exposure features that are explicitly learned to improve the nighttime semantic segmentation results. They also introduced one of the largest nighttime driving scene datasets called NightCity (Tan *et al.*, 2021). Building on the above work, (Deng *et al.*, 2022) proposed a dual-level architecture to cater to the large variations in the illumination in nighttime images. They proposed a regularized light adaptation module (ReLAM) that addresses the variation in illumination patterns while preserving the textures in nighttime images. It was aimed at better generalization for nighttime images by avoiding large domain shifts during adaptation.

To address the scarcity of nighttime driving datasets and annotations, (Wang *et al.*, 2022) proposed proprietary synthetic nighttime datasets and used the style transfer methodology. With the idea of effectively recognizing objects and their boundaries in dark environments, they proposed the semantic segmentation framework SFNET-N having a light enhancement network that could address the loss of feature information due to over/under exposure. However, the real-time performance of the model was not satisfactory, and there were issues with segmenting distant and small objects. (Song *et al.*, 2022) proposed an unsupervised convolutional neural network (CNN) based approach having an appearance transferring module and a segmentation module that are jointly trained end-to-end. The appearance transfer module was introduced to transfer the day and night images to a shared latent space or feature representation, thus exploring the matching semantic features in their content to improve the segmentation performance.

To our knowledge, there is no work specifically for the nighttime domain and nighttime urban driving scenarios in panoptic segmentation. There is also a scarcity of large nighttime driving datasets having panoptic annotations that are essential for deep learning. Considering the above works, we propose using a domain adaptation technique to address nighttime panoptic segmentation in urban driving scenarios by utilizing some nighttime driving datasets mentioned above. In their work, (Sun *et al.*, 2019) also demonstrated that the idea of converting nighttime images directly to daytime during inference does not produce effective results for the semantic segmentation task. However, in contrast, (Romera *et al.*, 2019) demonstrated that this idea could produce good results. Due to this ambiguity, we have opted not to adopt this approach for the panoptic segmentation task in this work, but it may be a good consideration for future research.

## V. METHODOLOGY

The method involves a few stages where, at first, a panoptic segmentation model is trained on a large daytime urban driving dataset having panoptic annotations. Then a domain adaptation model is trained, or a pre-trained model is used to translate a part of the original training dataset into the nighttime or poor illumination domain. This is followed by re-training the panoptic segmentation model on the adapted training dataset using the original daytime panoptic annotations. Since the same daytime panoptic annotations also apply to the converted nighttime images, the method helps to narrow down the performance gap on the nighttime images. Fig. 1 illustrates the high-level methodology used.

### A. Panoptic Segmentation model

Panoptic-DeepLab (Cheng *et al.*, 2020) is chosen as the deep-learning panoptic segmentation model architecture to be replicated for this study since it is a practical and efficient seminal architecture. ResNet (He *et al.*, 2016) is used as the backbone for feature extraction in this model.

### B. Domain adaptation model for day-to-night conversions

To reduce the domain gap between daytime and nighttime images, converting the daytime images with available panoptic annotations to the nighttime domain is the first step. For this purpose, CycleGAN (Zhu *et al.*, 2017) was utilized in this study with two different approaches. Firstly, in **Approach-1**, a pre-trained CycleGAN model available from (Nag, Adak and Das, 2019) was used for day-to-night conversion of urban driving images in the Cityscapes dataset.

**Robustness:** In order to try and improve the robustness of panoptic segmentation on a variety of lighting effects or illumination patterns seen in different nighttime urban driving environments, **Approach-2** was introduced. In this approach, either the CycleGAN was trained from scratch, or a trained model was refined with different combinations of nighttime datasets introduced earlier, such as Nighttime driving (Dai and Gool, 2018), NightCity (Tan *et al.*, 2021), Urban Night Driving Dataset (Nag, Adak and Das, 2019) and Dark Zurich (Sakaridis *et al.*, 2019) dataset, with the source dataset being Cityscapes (Cordts *et al.*, 2016) primarily.

### C. Generating nighttime images to enhance the training set

Cityscapes (Cordts *et al.*, 2016) dataset was chosen as the primary dataset for conversion to nighttime since it has publicly available panoptic annotations for the training and validation sets. (Sun *et al.*, 2019) demonstrated that converting ~28% of the training images from daytime to nighttime and using them for training produced the best semantic segmentation results for nighttime driving images. Motivated by this, we converted roughly ~28% (~840 images) of the Cityscapes training dataset to the nighttime domain using CycleGAN.

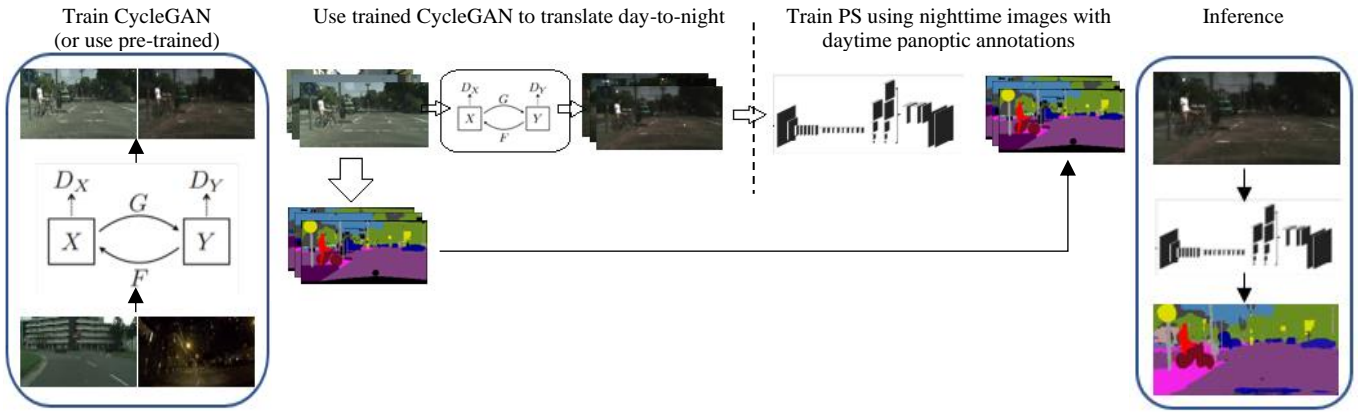

Fig. 1. The figure shows the high-level approach for this work. On the left, a domain adaptation model for day-to-night conversion is trained using unpaired images of day and nighttime. In the middle, daytime images with public panoptic annotations are converted to the nighttime domain using the trained domain adaptation model, and the panoptic segmentation model is trained using those nighttime images and the same daytime panoptic annotations. On the right, inference is performed on nighttime images based on the trained panoptic segmentation model. 'PS' refers to the Panoptic Segmentation model.

The converted images were used to update the Cityscapes training dataset, and the updated dataset was then fed to the panoptic segmentation model to retrain it.

### D. Evaluation methodology

**Quantitative evaluation:** Since there is a lack of large datasets with nighttime urban driving images with publicly available panoptic annotations, we decided to use the Cityscapes dataset for the quantitative evaluation. For this purpose, the Cityscapes validation set was fully converted to the nighttime or poor-illumination domain using the trained CycleGAN model(s). Subsequently, the quantitative evaluation could be performed using the original and converted nighttime validation sets.

**Qualitative evaluation** was performed using the daytime panoptic annotations in Cityscapes and visually on the nighttime dataset(s) not having the annotations.

## VI. EXPERIMENTS

### A. Datasets

**Cityscapes** (Cordts *et al.*, 2016) dataset contains 5000 images of street driving scenes with fine pixel-level panoptic annotations from 50 different cities. It primarily has daytime images with good/medium weather conditions for several months (summer, spring, fall). There are 2975 training, 500 validation and 1525 testing images with a resolution of 2048 x 1024. The annotations are divided into 19 major classes that describe a driving scene. We used this dataset for training and testing panoptic segmentation and used it as the primary dataset for training CycleGAN.

**Nighttime Driving** (Dai and Gool, 2018) dataset has 50 test images of nighttime driving with a resolution of 1920 x 1080. The semantic annotations used in this dataset have the same 19 classes as in the Cityscapes dataset described before. In our experiments, the dataset has been used to train the domain translation model (CycleGAN).

**Urban Night Driving Dataset (UNDD)** (Nag, Adak and Das, 2019) contains 7125 day and night images which are unlabelled. It also has 75 nighttime images that have pixel-level semantic annotations in the 19 Cityscapes evaluation classes. However, only the test set of 59 images with a resolution of 2048 x 1024 are publicly available in this dataset. The test set has been used for CycleGAN training and qualitative evaluation of the panoptic segmentation task.

**Dark Zurich** (Sakaridis *et al.*, 2019) dataset contains 8779 images of resolution 1920 x 1080 from the day, twilight and night conditions, along with their camera-based GPS coordinates as correspondences. It has 151 test and 50 validation images with pixel-level semantic annotations in 19 classes of Cityscapes. This dataset has been used for training CycleGAN and qualitative evaluation.

**NightCity** (Tan *et al.*, 2021) dataset contains 4297 real nighttime images with a resolution of 1024 x 512 with semantic annotations at the pixel level. This dataset has been used for training CycleGAN and qualitative evaluation.

**BDD100K** (Yu *et al.*, 2020) is a large driving dataset with 70K training, 10K validation and 20K testing images from various cities with a resolution of 1280 x 720. It has been used for qualitative evaluation in this work.

### B. Experimental settings

The training and evaluation for the panoptic segmentation model and CycleGAN model were conducted entirely on Google Colaboratory on a single GPU using Pytorch and Python-based libraries.

**Panoptic Segmentation** model Panoptic-DeepLab (Cheng *et al.*, 2020) was trained on the Cityscapes dataset. Similar to (Cheng *et al.*, 2020), the 'poly' learning rate schedule was used along with the Adam (Kingma and Ba, 2015) optimizer without weight decay. The number of training iterations was set to 90000, and we found the base learning rate of 0.00005 to be the best with this setting. Due to the memory constraints, the batch size was set to 1. In data-preprocessing, crop size of 1025 x 2049 (i.e., the whole image) was found to be the best, and it was used in the training and evaluation runs that have been reported. Unless otherwise specified, the ResNet-50 variant was used as the backbone in the reported results. Hard pixel mining with Cross Entropy loss was used for the semantic segmentation task with the loss weights set to 3 if a pixel belonged to an instance having less than 4096 area (or set to 1, otherwise). Mean Squared Error (MSE) was used for centre loss, and L1 loss was used for the offset prediction. Post-processing settings such as NMS (non-maximum suppression) kernel size (7), threshold (0.1) and k (200) values were kept the same as in the Panoptic-DeepLab (Cheng *et al.*, 2020) paper. More details about settings can be found in the supporting material.

**Domain translation in Approach-1** was performed via a pre-trained CycleGAN model from (Nag, Adak and Das, 2019) that was pre-trained on a mixture of nighttime images from BDD (Yu *et al.*, 2020), Mapillary Vistas (Neuhold *et al.*, 2017) and UNDD (Nag, Adak and Das, 2019) datasets.

**Domain translation in Approach-2** was performed by either training CycleGAN from scratch or refining a trained model. The model was trained for 200 epochs with a base learning rate of 0.0002, which was kept constant for the first 100 epochs and then decayed linearly to 0 in the next 100 epochs. Adam (Kingma and Ba, 2015) optimizer with a momentum of 0.5 was employed during the training. CycleGAN is very memory intensive due to four different sub-networks (two generators and two discriminators), and the Cityscapes dataset contains high-resolution images (i.e., 2048 x 1024). However, Google Colaboratory is a constrained memory environment. Because of the above reasons, a batch size of 1 along with a crop size of 256 x 256 was used for random cropping as a pre-processing step during training. Each training cycle took ~36 hours or more.

**Driving robustness via Approach-2**: we found that the nighttime images generated via Approach-1 were usually dark but had missing or fewer illumination patterns such as vehicle headlights/taillights, traffic lights, streetlights, building lights or reflections. This meant that a panoptic segmentation model trained on such images would not perform as robustly in varied nighttime urban driving scenes that contain such illumination patterns as in the NightCity, BDD100K or Dark Zurich (night) datasets. To address this concern, we decided to train CycleGAN with different combinations of nighttime datasets to induce better lighting effects in the generated nighttime images. TABLE *IV* in the Appendix shows a summary of the experiments conducted to train or refine different CycleGAN model(s) with the respective source and target datasets and training type (i.e., whether a model was refined or freshly trained).

In this work, we provide the results using the 'Solution-2' shown in TABLE *IV*. In 'Solution-2', CycleGAN is trained with the Cityscapes training set and a mixture of images from NightCity, Dark Zurich, Nighttime driving datasets and UNDD datasets (refer to TABLE *IV*, TABLE *V* and TABLE *VI* in the Appendix for more details). Then the retrained panoptic segmentation model from Approach-1 is refined for 90K iterations (i.e., 'Refined-1' with 30K iterations and further refined in 'Refined-2' with 60K iterations) using the generated images from the trained CycleGAN. Note that the ~28% method explained in section V.C applies here too.

**Translating images** When translating images from day to night using the trained CycleGAN model, the pre-processing step was skipped so that the full image resolution is taken as input without cropping.

*C. Evaluation metrics*

I have used the standard Panoptic Quality (PQ) (Kirillov *et al.*, 2019) and the related Segmentation Quality (SQ) and Recognition Quality (RQ) metrics for quantitative performance analysis of the panoptic segmentation model(s). For completeness, we have also reported mean Intersection-over-Union (mIoU), frequency weighted Intersection-over-Union (fwIoU), mean pixel accuracy (mACC), pixel accuracy (pACC), Average Precision (AP) and Average Precision-50 (AP50) metrics.

VII. RESULTS AND DISCUSSION

This section offers the quantitative and qualitative results for Approach-1 followed by Approach-2 and provides a discussion on the results.

*A. Using the pre-trained CycleGAN model (Approach 1)*

*a) Quantitative results:* TABLE I reports the experimental results for Approach-1 with different metrics. These results are obtained on the baseline and retrained Panoptic-DeepLab models, as tested on the original and the converted (nighttime or low illumination) Cityscapes validation set. All comparisons shown below are in absolute terms.

On the nighttime (converted) Cityscapes validation set, the retrained model performed much better than the baseline model and achieved a gain of +10.63%, +12.24% and +2.92% in PQ, RQ and SQ, respectively. The 'Stuff' category saw bigger improvement as compared to the 'Things' category with +11.81% (vs +9.03%), +13.4% (vs +10.76%) and +2.94 (vs +2.9%) in PQ, RQ and SQ respectively. Notably, an even bigger improvement was realized in the semantic segmentation sub-task with +14.43% in mIoU and +14.28% in mACC. The retrained model performed better in the instance-level semantic labelling metrics AP and AP50 on both the original and converted nighttime Cityscapes validation sets. However, on the nighttime validation set, the improvement in AP50 was significantly bigger (+10.07%) as compared to the daytime validation set (+2.19%).

Despite the significant improvements in the low-illumination or nighttime validation set images, the retrained model performed almost similar to the baseline model on the original (daytime) validation set, with only a loss of -0.63% (PQ), -0.57% (RQ) and -0.32% (SQ) as compared to the baseline model.

TABLE II reports the per-class PQ results for the 19 Cityscapes evaluation classes for the baseline and retrained models on the original cityscapes validation set and the converted nighttime / poor illumination validation set. For the nighttime validation set, the classes 'sky', 'train', 'sidewalk', 'truck' 'traffic light', 'pole', 'building' and 'traffic sign' attain double-digit performance improvement in the PQ metric with the retrained model (i.e. +30.3%, +20.4%, +19.8%, +17.0%, +15.2%, +12.9%, +12.5% and +12.0% respectively), as compared to the baseline model.

Even with these large per-class improvements on the nighttime (converted) dataset, the retrained model achieved an average PQ loss of only -0.6% over all the classes on the original daytime validation set.

Overall, these results demonstrate that, with Approach-1, the retrained model can attain significantly improved performance on the nighttime or low-illumination driving images while maintaining similar performance as the baseline model on the daytime images.

TABLE I. Comparison of quantitative results for panoptic segmentation on Cityscapes validation set(s) in Approach 1

| Model | Dataset | All | | | Things | | | Stuff | | | Semantic | | | | | |
|---|---|---|---|---|---|---|---|---|---|---|---|---|---|---|---|---|
| | | PQ (%) | SQ (%) | RQ (%) | PQ (%) | SQ (%) | RQ (%) | PQ (%) | SQ (%) | RQ (%) | mIoU (%) | fwIoU (%) | mACC (%) | pACC (%) | AP (%) | AP50 (%) |
| Baseline | Original | 53.78 | 78.91 | 66.62 | 41.55 | 76.65 | 53.94 | 62.68 | 80.55 | 75.85 | 73.24 | 91.36 | 81.67 | 95.31 | 20.13 | 39.1 |
| Retrained | Original | 53.15 | 78.59 | 66.05 | 41.29 | 76.49 | 53.74 | 61.78 | 80.12 | 75 | 72.11 | 91.21 | 80.94 | 95.22 | 21.89 | 41.29 |
| Baseline | Converted | 34.65 | 73.81 | 44.74 | 26.22 | 72.8 | 35.47 | 40.78 | 74.54 | 51.47 | 49.18 | 78.8 | 59.46 | 87.4 | 9.96 | 21.71 |
| **Retrained** | **Converted** | **45.28** | **76.73** | **57.01** | **35.25** | **75.7** | **46.23** | **52.59** | **77.48** | **64.87** | **63.61** | **86.77** | **73.74** | **92.59** | **16.06** | **31.78** |

'Baseline' model is trained with the original Cityscapes training set. 'Retrained' model is trained with updated Cityscapes training set having ~28% converted images. Both the models were trained with 90K iterations. 'Original' dataset represents Cityscapes original validation set. 'Converted' dataset represents the Cityscapes validation set converted to the nighttime or low-illumination domain. ResNet-50 is used as the backbone in both the 'Original' and 'Retrained' models.

TABLE II. The per-class Panoptic Quality (PQ) metrics on Cityscapes validation set(s) with original and retrained models in Approach 1

| Model | Dataset | Road | Sidewalk | Building | Wall | Fence | Pole | Traffic light | Traffic sign | Vegetation | Terrain | Sky | Person | Rider | Car | Truck | Bus | Train | Motorcycle | Bicycle |
|---|---|---|---|---|---|---|---|---|---|---|---|---|---|---|---|---|---|---|---|---|
| Baseline | Original | 97.0 | 73.0 | 87.2 | 28.9 | 33.5 | 51.5 | 44.5 | 66.2 | 88.5 | 33.5 | 85.7 | 43.4 | 36.9 | 58.0 | 40.0 | 56.6 | 31.8 | 31.0 | 34.7 |
| Retrained | Original | 97.2 | 71.9 | 87.3 | 24.7 | 32.3 | 51.3 | 45.3 | 66.3 | 88.4 | 29.9 | 85.1 | 41.7 | 37.3 | 56.9 | 38.4 | 55.0 | 36.9 | 30.6 | 33.6 |
| Baseline | Converted | 90.1 | 38.9 | 68.2 | 9.2 | 12.3 | 20.3 | 14.1 | 44.6 | 80.5 | 20.2 | 50.2 | 32.3 | 29.6 | 49.1 | 17.4 | 35.2 | 8.9 | 12.2 | 25.0 |
| **Retrained** | **Converted** | **94.2** | **58.8** | **80.7** | **17.4** | **21.4** | **33.2** | **29.3** | **56.6** | **84.3** | **22.1** | **80.5** | **36.0** | **33.0** | **53.7** | **34.3** | **48.5** | **29.3** | **20.3** | **26.9** |

*The same terminology as in TABLE I above is applicable here.

b) *Qualitative results:* Since the test set of the Cityscapes dataset does not have publicly available panoptic annotations, the validation set is used in the qualitative analysis. Fig. 2 and Fig. 3 provide the qualitative examination results on the Cityscapes converted validation set and UNDD test set.

In Fig. 2, in line with the per-class results examined in the quantitative analysis in part a) above, it can be seen that the retrained model achieved much higher accuracy for the sky, poles, traffic signs, sidewalk, and traffic lights. For example, the baseline model could not distinguish the sky in all the nighttime images, but the retrained model could. In rows 4, 8, 9 and 10, the traffic light poles are finely detected by the retrained model but not so by the baseline model. The traffic signs are more accurately segmented by the retrained model in rows 1, 4 and 7. In rows 5 and 6, the sidewalk is better segmented by the retrained model as compared to the baseline. Since the 'ground' category is not separately evaluated, it is shown as the sidewalk in rows 1, 2 and 3 as an example and visually better segmented by the retrained model. The buildings are segmented well in rows 2, 3, 5, 6, 8, 9 and 10 by the retrained model while being incorrectly merged with the sky by the baseline model. Rows 5 and 9 show improved traffic lights segmentation by the retrained model. It is observed that some classes like road, car, bicycle, and terrain have a lesser improvement in performance as the baseline model is also able to segment the corresponding objects well.

On a different note, in row 3, some reversed traffic signs are detected correctly by the retrained model but labelled as 'void' in the ground truth. Similarly, in row 4, two high wall pillars are predicted as buildings by the retrained model but labelled as 'void' in the ground truth. However, the baseline model cannot even detect and segment the reversed traffic signs and the wall pillars in the above two conditions. Thus, the retrained model is shown to perform better, even though a bit inaccurately due to less contrast, poor illumination and thus lack of clear object boundaries in the converted images.

**Problems:** We can also note some issues in the predicted results visually. The 'ego-car' region (source car dashboard area) is not segmented well by both the baseline and retrained models. This is a typical problem with panoptic segmentation for urban driving scenes, which should be separately addressed, for example, by (Chen *et al.*, 2020). In row 10, a bicycle is predicted as a motorcycle, although by both the models. This happens because the person is riding the bicycle in a more standup position, and the seat of the bicycle is higher, thus resembling a motorcycle more. In row 3, some poles are missing, and the bus is not recognized by both the baseline and retrained models. This happens because the converted nighttime images are very dark with no apparent illumination near the poles and none-to-very-low contrast in the boundaries of the bus, which results in the lack of detection of these as objects. One reason for this is that a high-quality domain translation of high-resolution images (2048 x 1024 in Cityscapes) is very difficult with CycleGAN because it is very memory intensive, and thus the input image must be cropped to a low resolution (i.e., 256 x 256 used in this work). It leads to blurry and noisy images when converted from day to night domain, which essentially blurs the object boundaries and creates difficulties in object detection and segmentation.

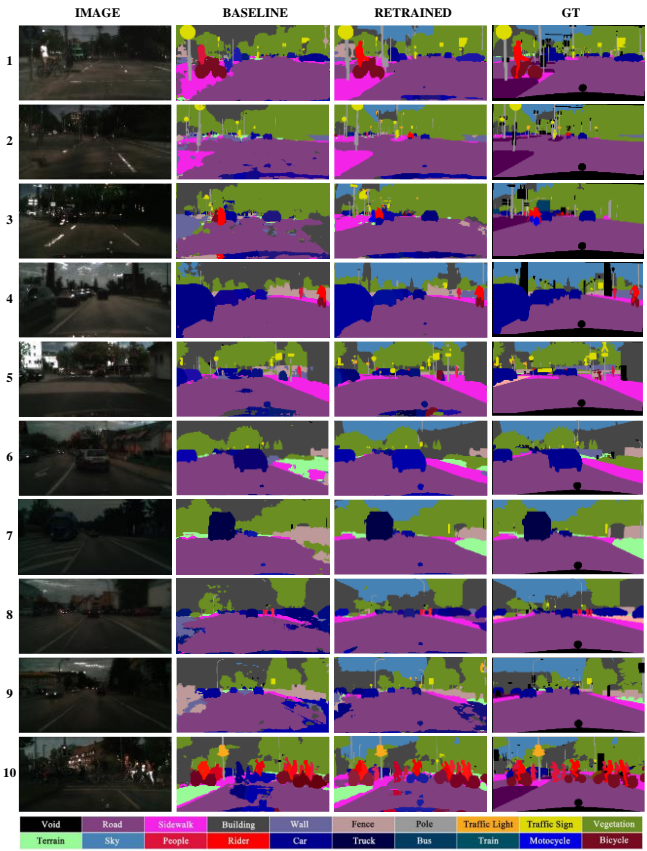

Fig. 2. Qualitative results on Cityscapes converted validation set showing the nighttime/poor illumination images, panoptic predictions with the baseline and retrained models (Approach-1) and the corresponding panoptic ground truth ('GT'). **This figure is best viewed in <u>high zoom</u> (or refer to Fig. 5 )**

One potential solution is to use a better backbone feature extractor or a higher variant of ResNet with more layers to extract better features. Another possibility is to use a super-resolution neural network to enhance the resolution of converted nighttime images, although this could increase the time taken to process an image.

In Fig. 3, we can notice better segmentation of poles, traffic lights, traffic signs and vehicles with the retrained model. This demonstrates that the retrained model that was trained on the converted Cityscapes training dataset can also perform better on another nighttime dataset, i.e., UNDD (test), when compared with the baseline model.

Qualitative evaluation was also performed with the other night driving datasets in VI.A. However, since the other datasets do not have a matching resolution with Cityscapes, on which the panoptic segmentation model was trained, the results were relatively poor for both models.

### B. Self-training CycleGAN to drive improved robustness in varied nighttime driving environments (Approach 2)

*a) Quantitative results:* TABLE III provides the quantitative metrics for Approach-2 (Solution-2) on the Cityscapes dataset. 'Refined-2' model is used in the analysis.

Compared with the results in TABLE I, although the 'Refined-2' model suffers about <3% on PQ, SQ, and RQ on the original dataset, it gains about +5% (PQ) and +6.3% (RQ) on the converted nighttime dataset generated from the pre-trained CycleGAN in Approach-1.

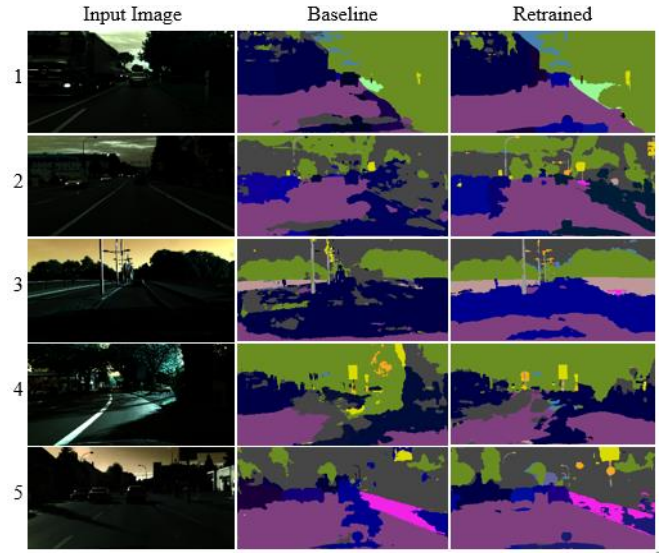

Fig. 3. Approach-1 Qualitative panoptic segmentation results on UNDD dataset using baseline and retrained models showing improved model predictions with the retrained model.

However, the performance gain is generally lesser than the Approach-1 retrained model. The metrics mIoU and mACC had an almost similar loss on the original dataset as the gain on the converted nighttime dataset, i.e., ~5%. In contrast, the instance-level semantic labelling metrics AP and AP50 showed a slightly higher loss on the original dataset than the gain on the converted dataset of Approach-1. However, the difference was only around ~2% at max.

These results show that the Refined-2 model is still effective on the converted nighttime validation set generated in Approach-1, though less so than the Approach-1 retrained model.

*b) Qualitative results:* Fig. 4 shows the predictions made on different nighttime driving datasets by three different models, i.e., baseline, Approach-1 retrained, and 'Refined-2' from Approach-2 Solution-2.

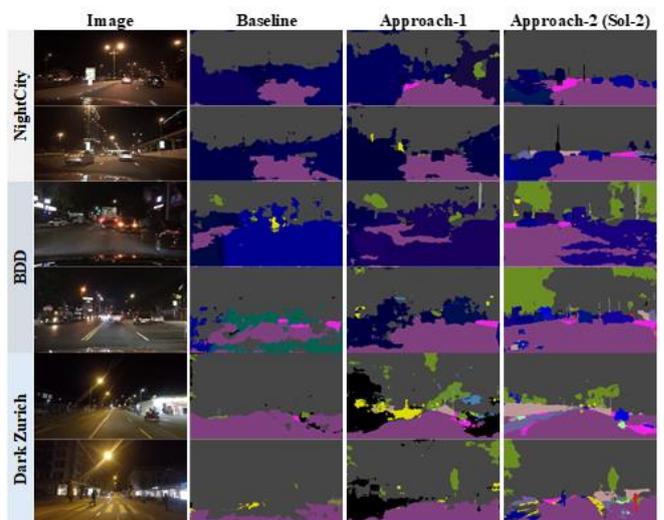

Fig. 4. Comparison of qualitative panoptic segmentation results on different nighttime datasets showing improved results with Approach-2 ('Refined-2' model from Solution-2). 'Approach-1' model refers to the retrained model in Approach-1. 'BDD' refers to BDD100K dataset.

TABLE III. Comparison of quantitative metrics for panoptic segmentation using Approach-2 (Solution-2) on Cityscapes validation set(s)

| Model | Dataset | Itr | All | | | Things | | | Stuff | | | Semantic | | | | | |
|---|---|---|---|---|---|---|---|---|---|---|---|---|---|---|---|---|---|
| | | | *PQ* (%) | *SQ* (%) | *RQ* (%) | *PQ* (%) | *SQ* (%) | *RQ* (%) | *PQ* (%) | *SQ* (%) | *RQ* (%) | *mIoU* (%) | *fwIoU* (%) | *mACC* (%) | *pACC* (%) | *AP* (%) | *AP50* (%) |
| Refined-1 | Original | 30K | 50.47 | 77.42 | 63.05 | 38.91 | 75.22 | 51.05 | 58.87 | 79.01 | 71.78 | 67.46 | 90.25 | 77.8 | 94.6 | 19.69 | 37.44 |
| Refined-2 | Original | 60K | 50.72 | 78.06 | 63.13 | 37.15 | 75.44 | 48.69 | 60.58 | 79.96 | 73.63 | 66.94 | 90.46 | 76.15 | 94.72 | 18.46 | 34.9 |
| Refined-2 | Converted | 60K | 39.76 | 75.19 | 51.097 | 29.46 | 74.29 | 39.49 | 47.26 | 75.86 | 59.54 | 54.12 | 81.72 | 65.01 | 88.99 | 12.46 | 25.86 |

Refined-1 model is refined from the 'retrained' model in Approach-1. Refined-2 model is refined further from the Refined-1 model. 'Itr' refers to the number of iterations used in the refinement training. 'Original' refers to the original Cityscapes validation set, 'Converted' refers to the Cityscapes validation set converted using Approach-1 pre-trained CycleGAN.

We can see that, even though the Approach-1 retrained model produced better results than the baseline model, the predictions are very bad visually. In contrast, the Approach-2 model (i.e., 'Refined-2') predictions are much better across different nighttime datasets, even though the panoptic segmentation model is not directly trained on any of them. For instance, a motorcycle, cars, sidewalks, and people are more accurately predicted by the Approach-2 model. It shows that the panoptic segmentation model trained through Approach-2 performs better in a variety of nighttime driving environments and thus has improved robustness.

**Problems:** However, the results also show that more improvements are needed to drive higher accuracy in predictions. For example, the areas with zero and very high brightness are labelled as 'building'. We can view this, for example, in the last two rows (i.e., Dark Zurich images), where high brightness traffic lights, the retail outlet and dark sky areas are predicted as buildings. One reason for such behaviour is the lack of zero illuminance sky regions in the generated nighttime images used for training the panoptic segmentation model. We should generate better nighttime images with a true nighttime sky to solve this issue. It could potentially be done using a twilight intermediate domain as used in some studies, such as (Dai and Gool, 2018) and (Nag, Adak and Das, 2019), or by more effective training of day-to-night domain adaptation models.

In summary, Approach-2 demonstrated that the robustness to varied nighttime driving conditions having different illumination patterns could be improved through a domain translation technique (at a low penalty to daytime performance).

## VIII. CONCLUSION

In this work, we presented two new methods for improving 1) performance and 2) robustness of deep learning-based panoptic segmentation on nighttime or low illumination urban driving scenes, using a domain adaptation technique. In Approach-1, we demonstrated that the use of the proposed domain translation technique led to a performance improvement of +10.63%, +12.24%, +2.92%, +14.43%, +14.28% and +10.07% in PQ, RQ, SQ, mIoU, mACC and AP50 respectively, on the converted Cityscapes nighttime validation set. It was also demonstrated that this improvement was achieved while maintaining the performance levels on daytime images. In addition, the work showed the respective improvement qualitatively on the UNDD dataset. While Approach-1 could not achieve robustness across multiple night-time driving datasets, we covered this gap in Approach-2. Approach-2 showed that it is possible to improve the robustness of panoptic segmentation to varied nighttime driving conditions with distinct illumination patterns by transferring knowledge through a domain translation approach without using such datasets to train the panoptic segmentation model directly.

A set of quantitative metrics were produced to support the argument(s). Predictions on nighttime images from BDD100K, NightCity and Dark Zurich datasets were produced to justify the improvement in robustness qualitatively. The work also highlighted some problem areas that need further attention and suggested related improvements.

## IX. FUTURE WORK

With the present work come certain problems and challenges, such as blurry and low-quality generated nighttime images, which affect the final performance. In future work, we could address such issues by using higher variants of ResNet or other backbone architectures, exploring super-resolution techniques to improve resolution and reduce noise, or by more efficient training of CycleGAN for high-resolution images. We can also investigate and evaluate the proposed methods using other large driving datasets like BDD100K as the primary dataset(s). In future, we need to identify methods to generate nighttime images with even better and more varied illumination patterns. We explored this via other Approach-2 solutions (i.e., other than solution-2). However, we can explore those solutions further to evaluate their impact on improving the panoptic segmentation task on nighttime driving images. Additionally, we can conduct studies to generate the true nighttime sky in translated images. I suggest that we could potentially achieve this by exploring the intermediate twilight domain, such as used by (Dai and Gool, 2018) and (Nag, Adak and Das, 2019). Also, we can explore the idea of conversion of nighttime images to daytime directly during inference, similar to the research by (Romera *et al.*, 2019) and (Sun *et al.*, 2019).

Taking inspiration from (Wu *et al.*, 2021) and (Deng *et al.*, 2022), there is also a possibility to embed new light adaptation modules within the Panoptic-DeepLab model architecture to improve nighttime performance. It should be explored further as an idea.

## ACKNOWLEDGEMENT

I would like to sincerely thank my supervisor Muhammad Sami Siddiqui for his continuous guidance and support during this project.

# Appendix

TABLE IV. Different experimental solutions for training or refining CycleGAN to generate more realistic nighttime driving images with better lighting effects (Approach-2)

| Solution | Source dataset | Target dataset(s) | Training Type | CycleGAN model |
|---|---|---|---|---|
| 1 | Converted Cityscapes training set from Approach-1 | Nighttime Driving (test) | Train | New |
| 2 | Cityscapes training set | MIX-1 | Train | New |
| 3 | Cityscapes training set | MIX-2 | Refine | Pretrained model |
| 4 | Cityscapes training set | Nighttime Driving (test) | Refine | Pretrained model |
| 5 | Dark Zurich - Twilight | Dark Zurich - Night | Refine | Trained model from solution 3 |

MIX-1 and MIX-2 datasets were constructed with images from NightCity, Dark Zurich, Nighttime Driving, and UNDD datasets as shown in *TABLE V*. Dark Zurich – Twilight and Dark Zurich – Night contained images twilight and night images from the Dark Zurich dataset respectively. Training type reflects whether the model was trained from scratch or it was refined based on a model from the 'CycleGAN model' column, where 'New' means trained from scratch, 'Pretrained model' refers to the pretrained day-to-night translation CycleGAN model from (Nag, Adak and Das, 2019) paper. Reasoning for the experiments is provided in TABLE *VI*.

TABLE V. Composition of MIX-1 and MIX-2 target datasets for CycleGAN training in Approach 2

| Dataset | Resolution | Total Images | MIX-1 | MIX-2 |
|---|---|---|---|---|
| Dark Zurich (train - night) | 1920 x 1080 | 2416 | 1167 | 937 |
| NightCity (train) | 1024 x 512 | 2998 | 1493 | 750 |
| UNDD (test) | 2048 x 1024 | 59 | 21 | 21 |
| Nighttime driving (test) | 1920 x 1080 | 50 | 50 | 50 |
| Total | | 5523 | 2731 | 1758 |

'train' and 'test' indicate training and test sets respectively. 'train – night' indicates nighttime images from the respective training dataset.

TABLE VI. Summary of purpose and reasoning for different experiments conducted in Approach 2

| Solution | Source dataset | Target dataset(s) | Type | CycleGAN model | Purpose / Action |
|---|---|---|---|---|---|
| 1 | Converted nighttime Cityscapes training set from Approach-1 | Nighttime Driving | Train | New | Attempt to induce the lighting effects from the Nighttime driving dataset |
| **2** | **Cityscapes training set** | **MIX-1** | **Train** | **New** | To induce varied nighttime lighting effects by using a combination of nighttime driving datasets and reduce the training time by filtering the target dataset images. |
| 3 | Cityscapes training set | MIX-2 | Refine | Pretrained model | To further train the pre-trained model to generate images that are more realistic in the night domain (having better streetlights/vehicle lights and building lights effects). Attempt to reduce the training time by filtering target dataset images. **Action:** Filtered the training datasets from MIX-1 further to reduce the number of images having colored reflection effects from the car dashboard |
| 4 | Cityscapes training set | Nighttime Driving | Refine | Pretrained model | Experimented with this solution based on the results of the Solution 1 as Solution 1 used converted images (not the real training set.) |
| 5 | Dark Zurich Twilight | Dark Zurich Night | Refine | Trained model from solution 3 | To try and refine Solution 3 model to be able to convert bright day time images or dusk time images to dark nighttime images as Dark Zurich has twilight and night images which can be used for training. **Action:** Used the 60th epoch model from solution 3 for further refinement as it produced the best results in the conversion tests. |

*Terminology used is same as in TABLE *IV*. Solution 2 has been used for reporting results in this work.

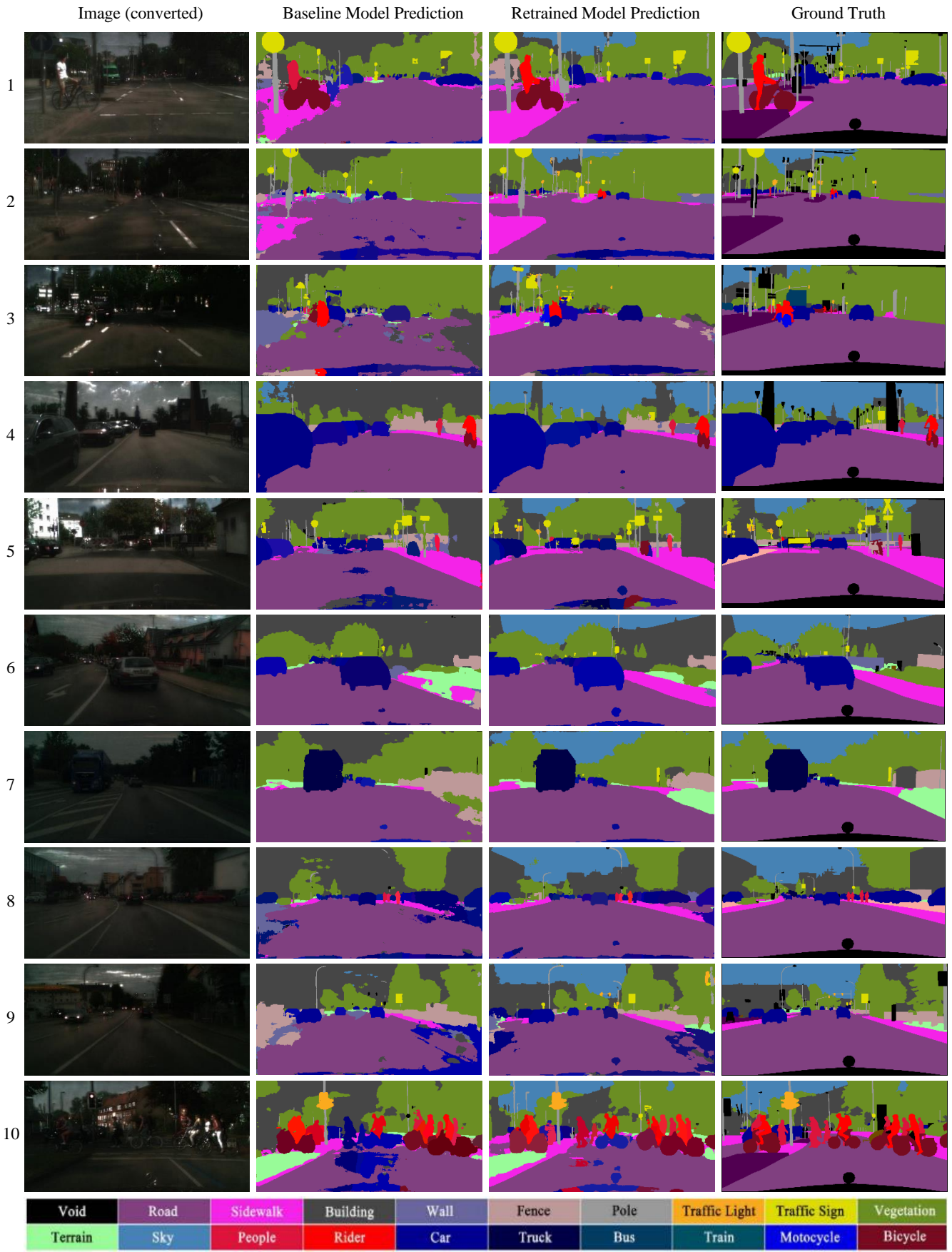

Fig. 5. Approach-1 qualitative results on the converted nighttime Cityscapes validation set with a high zoom factor.

| Real Image | Converted Image (with bright spots) |

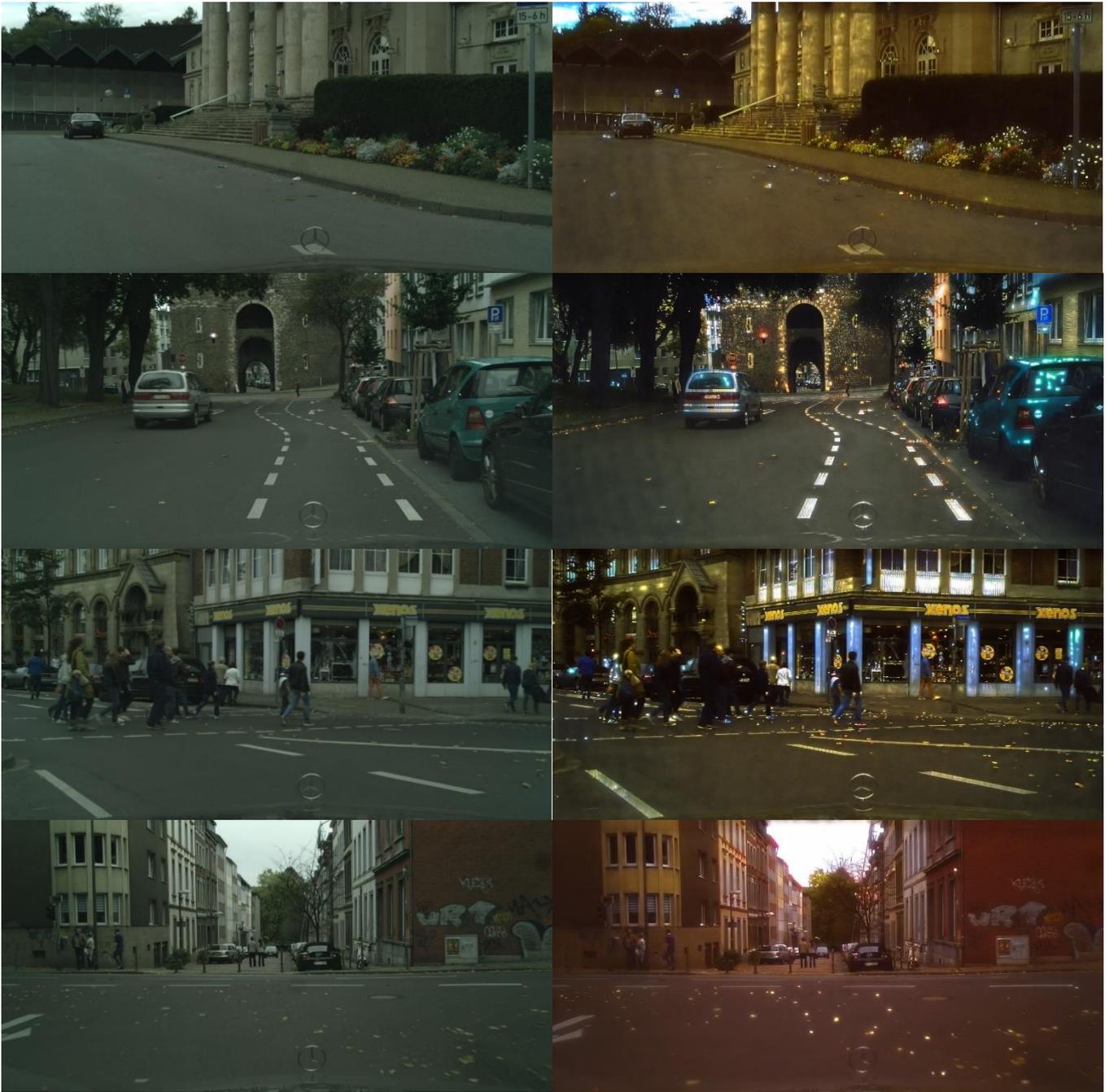

Fig. 6. Conversion of images leading to bright spots in small high-contrast areas of the image, including leaves, buildings, car windows.